\documentclass[letterpaper, 10 pt, conference]{ieeeconf}  % Comment this line out if you need a4paper

\IEEEoverridecommandlockouts                              % This command is only needed if 
\usepackage{graphics} % for pdf, bitmapped graphics files
\usepackage{graphicx}
\usepackage{times} % assumes new font selection scheme installed
\usepackage{amsmath} % assumes amsmath package installed
\usepackage{amssymb}  % assumes amsmath package installed
\usepackage{mathrsfs}
\usepackage{marvosym}
\usepackage{amsfonts}
\usepackage{kantlipsum}

\makeatletter
\let\NAT@parse\undefined
\makeatother
\usepackage[numbers,sort&compress]{natbib}

\usepackage{hyperref}
\usepackage{graphicx}
\usepackage{float}
\usepackage{ctable}
\usepackage{cuted}
\usepackage{colortbl}
\usepackage{multirow}
\let\labelindent\relax
\usepackage{enumitem}

\usepackage{amsmath}
\usepackage{algpseudocode}% http://ctan.org/pkg/algorithmicx
\usepackage{algorithm}% http://ctan.org/pkg/algorithm

\usepackage{graphicx}
\usepackage{float}
\usepackage{ctable}
\usepackage{cuted}
\usepackage{colortbl}
\usepackage{multirow}
\usepackage{threeparttable}
\usepackage{makecell}
\usepackage{wrapfig}

\definecolor{green}{RGB}{0,150,10}
\definecolor{blue}{RGB}{0,148,181}
\definecolor{orange}{RGB}{194,153,107}

\usepackage{amssymb}

\newcommand{\methodname}{SKT}
\newcommand{\APAll}[0]{mAP\textsubscript{2,4,8}\;}
\newcommand{\APAllArrow}[0]{\APAll($\uparrow$)}

\usepackage{pifont}
\newcommand{\xmark}{\ding{53}}%

\newcommand{\AKDArrow}[0]{AKD($\downarrow$)}
\title{\LARGE \bf
SKT: Integrating State-Aware Keypoint Trajectories with Vision-Language Models for Robotic Garment Manipulation
}
\author{Xin Li$^{1*}$, Siyuan Huang$^{1,6*}$, Qiaojun Yu$^{1}$, Zhengkai Jiang$^{2}$, \\ Ce Hao$^{3}$, Yimeng Zhu$^{4}$, Hongsheng Li$^{5}$, Peng Gao$^{6}$\textsuperscript{\Letter} and Cewu Lu$^{1}$\textsuperscript{\Letter}% <-this % stops a space
\thanks{$^{1}$Xin Li, Siyuan Huang,  Qiaojun Yu, Cewu Lu are with the Shanghai Jiao Tong University, China. $^2$Zhengkai Jiang is with The Hong Kong University of Science and Technology, HongKong. $^3$Ce Hao is with Department of Computer Science, National University of Singapore, Singapore. $^{4}$Yimeng Zhu is with the Yuandao AI. $^{5}$Hongsheng Li is with CUHK-MMLab, Hong Kong. $^6$Siyuan Huang and Peng Gao are with the Shanghai AI Lab. * indicates an equal contribution. \Letter Peng Gao and Cewu Lu are the equal corresponding authors, \texttt{gaopeng@pjlab.org.cn, lucewu@sjtu.edu.cn}}
}
\begin{document}

\maketitle
\thispagestyle{empty}
\pagestyle{empty}

\begin{abstract}

Automating garment manipulation poses a significant challenge for assistive robotics due to the diverse and deformable nature of garments. Traditional approaches typically require separate models for each garment type, which limits scalability and adaptability. In contrast, this paper presents a unified approach using vision-language models (VLMs) to improve keypoint prediction across various garment categories. By interpreting both visual and semantic information, our model enables robots to manage different garment states with a single model. We created a large-scale synthetic dataset using advanced simulation techniques, allowing scalable training without extensive real-world data. Experimental results indicate that the VLM-based method significantly enhances keypoint detection accuracy and task success rates, providing a more flexible and general solution for robotic garment manipulation. In addition, this research also underscores the potential of VLMs to unify various garment manipulation tasks within a single framework, paving the way for broader applications in home automation and assistive robotics for future. 
The project page is available at \href{https://sites.google.com/view/keypoint-garment/home}{{\texttt{sites.google.com/view/keypoint-garment}}}.

\end{abstract}
\section{Introduction}

Garments, as one of the most ubiquitous items in home environments, have long been a focal point in assistive robotics research~\cite{avigal2022speedfolding,canberk2023clothfunnels}. Tasks such as washing, folding, and ironing garments exemplify how robots can assist with everyday household activities. However, despite advancements in robotic vision and manipulation technologies, accurately recognizing and manipulating garments remains a challenge due to their diverse shapes and deformability~\cite{longhini2024unfolding,bertiche2020cloth3d}. Robots must not only recognize keypoints on garments but also adapt to the constantly changing states of these flexible objects to perform precise manipulations. This adaptability is essential for ensuring reliable performance in complex real-world scenarios.

Traditional garment manipulation methods often depend on 3D data and class-specific keypoint recognition models~\cite{lips2024learning,wu2024unigarmentmanip,ralclothpoint}, which are inherently limited by their inability to generalize across various garment types or states. These models typically perform well only in narrow contexts, struggling to infer or adapt to unknown garment configurations or unstructured states. For instance, a model trained to recognize keypoints on flat garments may fail when faced with a crumpled or folded item, limiting its effectiveness in dynamic environments. These limitations underscore the need for a more flexible and scalable approach capable of handling a wide range of garment states without requiring separate models for each garment category.
\begin{figure}[t]
    \centering
    \vspace{-4mm}\includegraphics[width=1\linewidth]{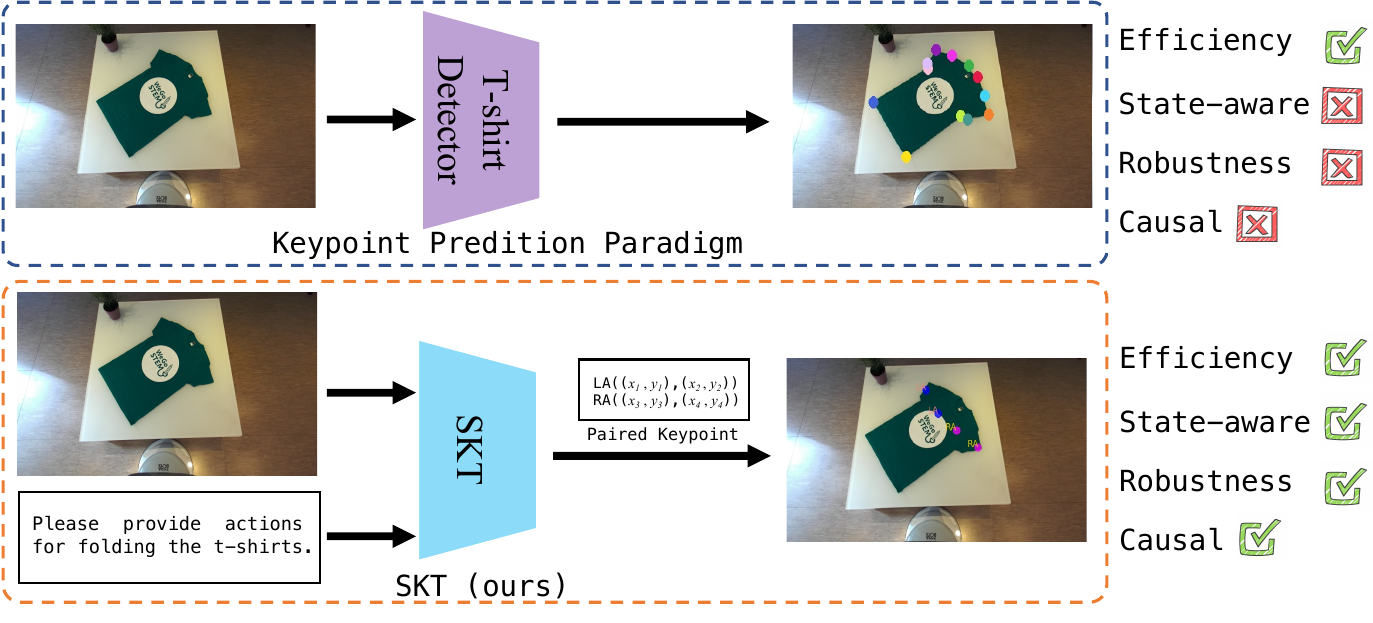}
    \caption{Comparison of Keypoint Detection Methods. The previous method~\cite{lips2024learning} struggles with deformed or ambiguous garment states, leading to inconsistent and incomplete keypoint predictions. In contrast, our \methodname~ utilizing state-aware paired keypoints and vision-language models (VLMs), achieves more robust and accurate keypoint detection, improving generalization across flat, folded, and deformed garment configurations.}
    \label{Fig: intro}
    \vspace{-0.4cm}
\end{figure}

To address these limitations, we propose state-aware paired keypoint formation, which generates \textbf{S}tate-aware \textbf{K}eypoint \textbf{T}rajectories for vision-language models (\methodname). By utilizing the unified paired keypoints trajectory formulation, our proposed method generalizes well to various garment environments, such as flat, folded, and deformed garments (Fig.~\ref{Fig: intro}). Furthermore, by harnessing the combined power of vision and language, our approach enables robots to interpret visual cues alongside textual descriptions of garment parts and manipulation tasks. the system to go beyond traditional 3D data, providing a more holistic understanding of the garment's current state and its corresponding keypoints. This integration enables the system to transcend traditional 3D data, offering a more holistic understanding of the garment's current state and its corresponding keypoints. Finally, the use of VLMs enables the robot to dynamically adapt to various garment states, as the model can process both visual features and language-based queries related to the manipulation tasks at hand.

To train our model, we created a synthetic dataset that covers a wide range of garment configurations, including flat, deformed, and folded states. By leveraging advanced physics simulators and rendering technologies, we simulated realistic garment deformations that represent the conditions encountered during robotic manipulation tasks. The synthetic dataset enhances scalability by eliminating the need for labor-intensive real-world data collection, enabling the model to generalize across a broad spectrum of garment types and configurations. Furthermore, by simulating garment deformations and creating associated text queries, we train the robot to predict keypoints using both visual and semantic information. This improves the robot’s ability to perform manipulation tasks, such as folding or rearranging garments, with greater precision and flexibility.

Notably, a key aspect of our approach is the introduction of reasoning-based vision-language tasks. It further optimize keypoint trajectories by enabling the robot to reason about garment states and adjust its actions accordingly. For example, if a garment is partially folded or deformed, the robot can infer the most relevant keypoints from the visual context and the provided semantic descriptions. This level of reasoning is essential for managing the complexities associated with deformable objects such as garments.

The main contributions of this study include: 
\begin{itemize} 
\item A unified paired keypoint trajectories formulation that integrates vision-language models to enhance robotic manipulation of garments. By combining visual and semantic information, the method enables robots to adapt to a wide variety of garment states and configurations. 

\item The creation of a large-scale synthetic dataset covers diverse garment states, improving the robot’s ability to generalize across various types of garments and manipulation scenarios.

\item We propose reasoning-based vision-language tasks that further improve keypoint detection by enabling the robot to infer and adjust to changing garment states, enhancing precision and adaptability in real-world applications.
\end{itemize}

\vspace{-0.1cm}
\section{Related Work} \label{Sec: related works}
\vspace{-0.1cm}

\subsection{Robotic Garment Manipulation}
\vspace{-0.05cm}
Robotic manipulation of garments is a critical challenge in assistive robotics due to the deformable and highly variable nature of clothing items~\cite{zhu2022deformablereview1,doumanoglou2016folding,avigal2022speedfolding,ha2022flingbot,wu2024unigarmentmanip}. Much of the existing research has focused on key tasks such as unfolding (flattening) and folding garments. While unfolding systems have made notable progress~\cite{canberk2023clothfunnels,proesmans2023unfoldir,wu2024unigarmentmanip}, they often leave garments imperfectly flattened, and their ability to handle a wide variety of clothing types and environmental conditions remains limited.

Traditionally, folding tasks for flattened garments rely on predefined state representations and scripted policies~\cite{canberk2023clothfunnels,de2022effective}. Researchers have explored various approaches for generating these state representations, including template fitting~\cite{avigal2022speedfolding,canberk2023clothfunnels} and semantic keypoint detection~\cite{lips2022learning,wu2024unigarmentmanip,corona2018clothpoints,seita2019keypoints}. Many of these methods leverage depth images~\cite{corona2018clothpoints,qian2020clothsegmentation,seita2019keypoints}, which are advantageous for capturing geometric data unaffected by lighting or background clutter. However, depth images can omit critical visual information such as garment patterns and seams, which can be important for accurate manipulation~\cite{qian2020clothsegmentation}. To address these limitations, our work utilizes RGB images, which offer richer visual information that can be crucial for precise garment manipulation.

% In this study, we focus on detecting keypoints on different states garments resting on a surface, which allows us to develop more robust systems for downstream manipulation tasks, such as folding. Our method is designed to generalize beyond specific garment types, enabling the robot to work with multiple clothing items and environmental setups by building on the output from existing unfolding systems.
\vspace{-0.05cm}
\subsection{Synthetic Data for Robotic Garment Manipulation}
\vspace{-0.05cm}
The use of synthetic data has become increasingly prevalent in robotic cloth manipulation, particularly for training models that need to generalize across diverse garment configurations and manipulation tasks~\cite{matas2018sim2real,corona2018clothpoints,seita2020smoothing,canberk2023clothfunnels,lips2024learning}. However, creating high-quality 3D assets that represent a wide variety of garments and states remains a significant challenge. Some approaches rely on limited sets of manually annotated pre-made cloth meshes~\cite{corona2018clothpoints,canberk2023clothfunnels,ganapathi2021DON}, while others use procedural generation techniques to produce single-layer meshes~\cite{matas2018sim2real,lips2024learning}. Although large-scale datasets like Cloth3D~\cite{bertiche2020cloth3d} have been developed, they often lack detailed semantic annotations required for precise manipulation tasks.

In this work, we extend the use of synthetic data to improve keypoint detection in garment manipulation tasks. Our pipeline generates diverse garment configurations, including varying deformations, allowing us to train models that can adapt to different garment types, states, and environmental conditions. By leveraging this approach, we aim to enhance the robot’s ability to recognize key manipulation points on garments, enabling more efficient and accurate operations in real-world scenarios.

\vspace{-0.05cm}
\subsection{Dense Representations for Garment Manipulation}
\vspace{-0.05cm}
Dense object descriptors, which capture point- or pixel-level object representations, have been widely applied to various robotic manipulation tasks~\cite{florence2018dense}. These descriptors have been extended in numerous works to propose grasping poses~\cite{simeonov2022neural, yen2022nerf}, manipulate deformable objects like ropes~\cite{sundaresan2020learning}, and smooth fabrics~\cite{ganapathi2021learning}. Additionally, point-level affordance learning has been explored for articulated objects~\cite{ling2024articulated,li2024unidoormanip}, deformable objects~\cite{zhang2020clothkeypoints}, and even in tasks involving language-guided~\cite{xu2024naturalvlm} and bimanual manipulation~\cite{wu2024unigarmentmanip,zhao2022dualafford}. These approaches enable more effective interaction and contact point selection, facilitating a variety of downstream tasks~\cite{wu2024unigarmentmanip,seita2021learning}. 
In this paper, our work extends the use of dense correspondence by applying these dense point representations specifically to deformable garment manipulation. By leveraging point-level affordance learning, we aim to enhance the robot’s ability to detect manipulation-relevant keypoints and improve performance across diverse garment states.
\begin{figure*}[t]
    \centering
    \vspace{-0.1cm}
    \includegraphics[width=1\linewidth]{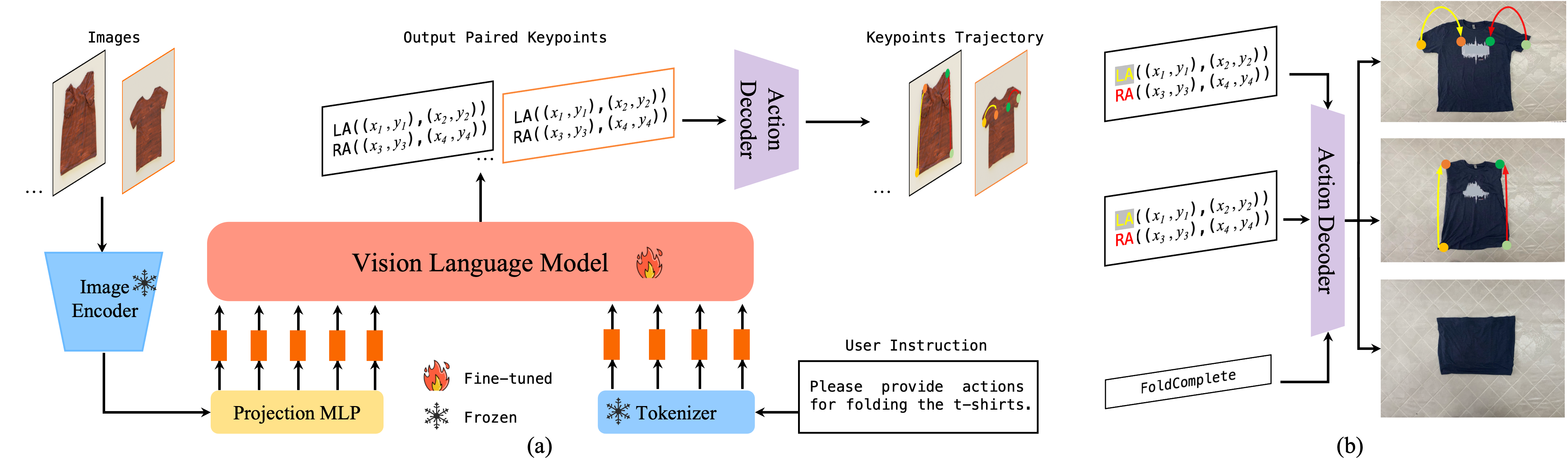}
    \caption{(a)The \textbf{overall framework} of state-aware keypoint trajectory (SKT). SKT generates action trajectories for clothes manipulation by leveraging a fine-tuned vision-language model for state-aware paired keypoint and action generation through the action decoder (b). }
    \vspace{-0.4cm}
    \label{Fig: architecture}
\end{figure*}

\vspace{-0.1cm}
\section{Method}
\vspace{-0.1cm}
In this section, we provide a detailed introduction to our proposed vision language keypoint prediction method for robotic garment manipulation. Our approach is based on the integration of vision-language models to enhance the robot’s ability to recognize keypoints on garments in different states in a unimodel, improving both flexibility and accuracy during manipulation tasks. To achieve this, we have developed a synthetic garment dataset and designed a keypoint prediction framework, which are described in the following subsections.
\vspace{-0.05cm}
\subsection{Synthetic Dataset Generation}
\vspace{-0.05cm}
To simulate the diverse and deformable nature of garments, we developed a synthetic dataset that encompasses a broad spectrum of garment configurations, ranging from flat to deformed and folded states. By leveraging advanced physics simulation tools like Blender, we modeled various household garments such as shirts, shorts, and towels. The dataset was structured to capture different deformation stages by applying realistic physical forces, thus simulating the conditions typically encountered during robotic manipulation tasks. The challenge of working with deformable objects required not only generating cloth meshes but also simulating a range of possible configurations, ensuring a broad distribution of garment deformations. The following sections provide a detailed explanation of the processes involved in mesh generation, deformation simulation, and the generation of corresponding images and keypoint annotations.

\subsubsection{\textbf{Garment Mesh Generation}}

The initial step involves the creation of garment meshes in multiple folded states. We began by defining a set of 2D boundary vertices based on templates specific to each garment type, inspired by~\cite{lips2022learning}. These vertices were connected using Bézier curves to form single-layer meshes, enhancing realism by representing features such as the neckline of a T-shirt with smooth edges. Parameters for the mesh skeleton, Bézier curves, and corner rounding radii were sampled from carefully calibrated ranges to ensure variability. The meshes were then triangulated with edge lengths constrained to 1 cm, and UV maps were generated to allow for subsequent texturing. During the folding process, we tracked vertices corresponding to key semantic regions, enabling automatic keypoint labeling at different folding stages.

\subsubsection{\textbf{Garment Mesh Deformation}}

Using Blender's physics simulation capabilities, we deformed the generated garment meshes by applying randomized orientations and allowing them to fall naturally to create wrinkles. Additionally, we simulated folding motions by performing circular grasping actions. The cloth was also lifted and rotated to produce both visible and hidden folds, with the focus on generating realistic fold patterns rather than fully crumpled states. Key physics properties such as bending stiffness, stretching, friction, and drag were randomized to increase diversity, and parameters were fine-tuned to maintain physical plausibility.

\subsubsection{\textbf{Garment Image Generation}}

\begin{figure}[t]
    \centering
    \includegraphics[width=0.95\linewidth]{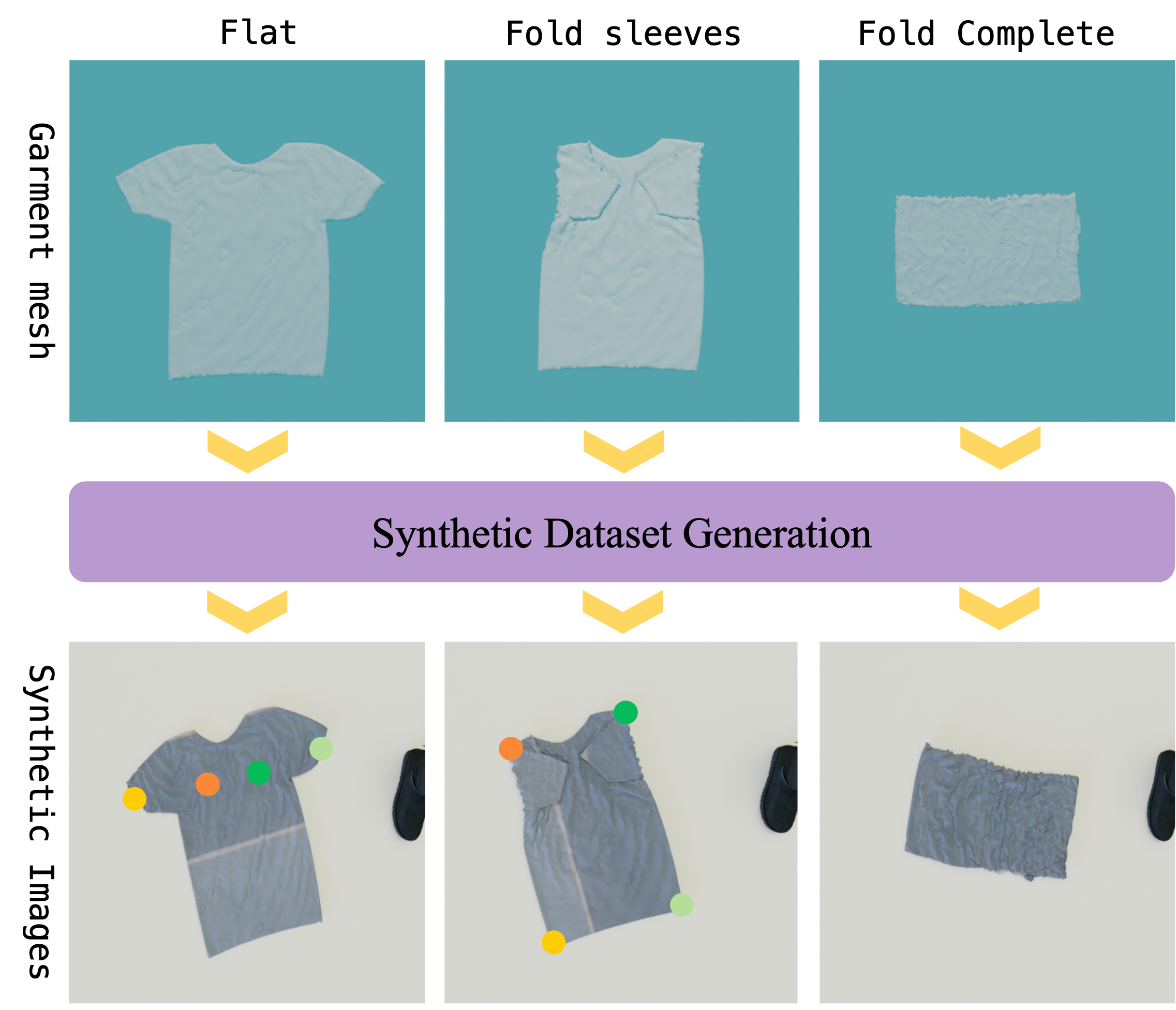}
    \vspace{-0.1cm}
    \caption{A sample set comprises synthetic images depicting different fold states with corresponding paired action keypoints annotations.}
    \label{Fig: imagesample}
    \vspace{-0.2cm}
\end{figure}

To generate high-quality synthetic images, we applied textures sourced from PolyHaven~\cite{polyhaven2023poly} to both the environment and the folding surface. Cloth meshes were solidified and textured, and distractor objects from the Google Scanned Objects dataset~\cite{downs2022google} were strategically placed in the scene to improve keypoint detection in varied environments. Cameras were positioned randomly around the garment, and the Cycles rendering engine was employed to produce photorealistic images with a range of lighting and viewpoint variations. A few examples of the synthetic image
can be found in Fig. ~\ref{Fig: imagesample}.
\vspace{-0.05cm}
\subsubsection{\textbf{Keypoint Generation}}
\vspace{-0.05cm}
For keypoint generation, we employed a raycasting technique to verify the visibility of vertices corresponding to each keypoint. Following the methodology in~\cite{lips2024learning}, we aimed to align synthetic labels with human annotations, recognizing that human-labeled keypoints may not always precisely match the ground truth. A keypoint was considered visible if any vertex within its 2-ring neighborhood was visible in the rendered image. Our experiments demonstrated that this approach better replicated human annotation patterns, resulting in more accurate and robust synthetic labels.
\vspace{-0.05cm}
\subsection{Paired Keypoint Representation}
\vspace{-0.05cm}
In garment manipulation, a set of keypoints—potentially represented as a nested structure, such as a skeleton—can effectively capture the state of a garment, regardless of its condition (e.g., degrees of wrinkling) or original shapes (including sizes and styles)~\cite{wu2024unigarmentmanip}. Compared to general representation formats used in MLLMs, such as bounding boxes, which may lack visual constraints in certain positions~\cite{gao2024sphinx}, keypoints retain critical structural information. This makes them more effective for capturing the nuances of garment shapes and configurations, as illustrated in Fig. 1.  Moreover, downstream policies primarily focus on the operational relationships between corresponding keypoints—essentially aligning Point A to Point B. By utilizing keypoint tuples for representation, we ensure compatibility with these existing policies, facilitating seamless integration into various manipulation strategies. To guide our model in learning the keypoint representation, we use the keypoints generated in the previous steps and formulate keypoint detection as a Visual Question Answering (VQA) task.

\vspace{-0.05cm}
\subsection{Action Tuple Trajectory Generation}
\vspace{-0.05cm}
\label{subsec:action_tuple}
Based on the keypoint representation proposed before, we can frame the task of garment folding action prediction as first implicitly identifying a set if keypoint coordinates tuple \(\{(x_1, y_1), (x_2, y_2), (x_3, y_3), (x_4, y_4)\}\) that correspond to the optimal grasping points for folding, based on the garment’s state. These keypoints are then formulated into action tuples, such as \(LA((x_1^l, y_1^l), (x_2^l, y_2^l))\) and \(RA((x_1^r, y_1^r), (x_2^r, y_2^r))\), where  LA and RA represent the "left arm" and "right arm," respectively, highlighting our method's capability to support bi-arm manipulation. This tuple representation not only links points, providing more structured information than discrete keypoints, but also ensures compatibility with robotic action primitives. By focusing on these key coordinates, the model can directly translate them into manipulation actions like grasping and folding, thereby enhancing precision. After obtaining the paired action point, our action decoder generates an action trajectory conditioned on the action primitive. The action model
is based on manually designed rules. Furthermore, the proposed SKT is adaptable to various garment types and folding scenarios, effectively guiding robotic movements and identifying key contact points. Training with multi-task data further enhances the model's generalization across diverse robotic manipulation tasks.

\vspace{-0.05cm}
\subsection{Vision Language Model Fine-Tuning}
\vspace{-0.05cm}
\subsubsection{\textbf{Model Architecture}} As shown in ~\ref{Fig: architecture},
 our vision language model is built upon the SPHINX-X framework~\cite{gao2024sphinx}, with LLaMA2 serving as its core language backbone. We chose this model due to its unique capability to concentrate on fine-grained, region-specific details of objects, making it well-suited for tasks requiring detailed visual analysis. Our model employs the "any resolution" strategy introduced by SPHINX~\cite{gao2024sphinx}. The input images are divided into smaller sub-images, after which the visual encoders process them to extract essential features. Given the dual need for both global and local visual comprehension in manipulation tasks, we integrate several image encoders: CLIP~\cite{radford2021learning} and DINOv2~\cite{oquab2023dinov2} to capture localized semantic information, and QFormer ~\cite{li2023blip} to summarize global features. These local and global features are then concatenated at the channel level. The spatial alignment between visual tokens and their corresponding language tokens is managed through projection layers. 

\subsubsection{\textbf{Fine-tuning Strategy}}
Our SKT training approach follows standard VQA methodology, embedding garment understanding details into natural language structures. To address visual domain gaps between our specialized dataset and generic imagery, as well as task representation gaps between action tuples and general image descriptions, we implement a two-phase fine-tuning approach. In the initial phase, we fine-tune visual projection layers using simple keypoint detection tasks. Templates like "Please detect the keypoints used to manipulate the [garment-type]" are employed. Outputs are keypoints of the garment, prefixed with "$<kp>$" to indicate their nature and facilitate post-processing. The second phase involves simultaneous fine-tuning of visual projection layers and the language model using an instruction-following dataset. We use templates like "Please provide actions for folding the [garment-type]", with the MLLM generating action tuples as described in Sec.~\ref{subsec:action_tuple}. These outputs are prefixed with "$<action>$" for clarity and ease of extraction. This staged approach gradually guides the MLLM to first understand garment keypoints, then link these point tuples with specific actions and instructions. In the second stage, we train the MLLM on both keypoint detection and action tuple generation tasks, reducing the ratio of keypoint detection tasks.

\vspace{-0.1cm}
\section{Experiments}
\vspace{-0.1cm}
\begin{table*}[t]
    \centering
    \vspace{-0.1cm}
    \caption{Resuls on aRTF dataset.}
    \vspace{-0.1cm}
    \label{tab:main-results}
    \begin{threeparttable}
    \begin{tabular}{lcccccccc}
        \toprule
       & &   & \multicolumn{3}{c}{\textbf{\APAllArrow}} & \multicolumn{3}{c}{\textbf{\AKDArrow}} \\
        \cmidrule(lr){4-6}
        \cmidrule(lr){7-9}
   \textbf{Data Source}  &      \textbf{Type}  &  \textbf{Methods} & T-shirt & Shorts  & Towel   & T-shirt & Shorts  & Towel  \\ 
        \midrule
          \multirow{4}{*}{Sim-to-Real} 
       & \multirow{3}{*}{Type-Specific~\cite{lips2024learning}} 
  & T-shirt Detector & 58.2 & - &- &14.0&- &- \\ 
    &          & Shorts Detector & - &51.4 &-&-  &27.3&- \\ 
 
  &     & Towel Detector & -  & -& 83.2 & - & - & 13.1 \\ 
       \cmidrule(lr){2-9}
 & { Uni-model }& SKT (ours) & 63.3& 56.7  & 83.9 & 8.7 &  10.9&  3.4 \\ 
  \midrule\midrule
      \multirow{5}{*}{Sim+Real-to-Real} 
       & \multirow{3}{*}{Type-Specific~\cite{lips2024learning}} 
  & T-shirt Detector & 69.1 & - &- &8.3&- &- \\ 
    &          & Shorts Detector & - &64.9 &-&-  &11.2  &- \\ 
 
  &     & Towel Detector & -  & -& 88.6 & - & - & 6.8  \\ 
       \cmidrule(lr){2-9}
 & \multirow{2}{*}{Uni-model} & SKT\_KP (ours) & 71.7& 60.3  & 88.2 & 6.8 &  7.9&  2.0  \\ 
 &   & SKT (ours) & 66.8& 59.9 & 86.8& 8.1 &  7.1&  3.0  \\ 
\bottomrule
    \end{tabular}
\begin{tablenotes}
\item Comparison with previous methods under different training data settings. The table compares the performance of type-specific models and the unified SKT model in Sim-to-Real and Sim+Real-to-Real training settings.
\end{tablenotes}
\end{threeparttable}
\vspace{-4mm}
\end{table*}

\subsection{Experimental Settings}
\vspace{-0.05cm}
To train and evaluate the effectiveness of the proposed method \methodname, we not only used synthetic data but also introduced the aRTF clothing dataset~\cite{lips2024learning}. aRTF dataset is collected from 14 real-world domestic settings, each characterized by unique environmental features and a variety of clothing items. It is meticulously segmented into 6 training scenes and 8 testing scenes, encompassing 15 towels and T-shirts within the training set, and an increased quantity of 20 in the test set. Similarly, the dataset comprises 8 shorts for training and 9 for testing. Notably, for each category, the dataset offers 210 images for training and an expanded set of 400 images for testing, with the exception of shorts, which are represented by 112 training images and 180 test images. More formally, we train the SKT model on a syntehtic dataset of 20,000 images and integrate the real-world data by employing fine-tuning techniques on the aRTF training dataset, thereby enhancing SKT's adaptability and accuracy across various garment types, including T-shirts, shorts, and towels.

We fine-tuned the SKT within the SPHINX framework~\cite{gao2024sphinx} and utilized eight NVIDIA A100 GPUs, each equipped with 80 GB. The fine-tuning process was completed over 3 epochs, with a total runtime of approximately 4 hours. The
visual encoders were kept frozen throughout the fine-tuning stage to preserve the pre-trained feature quality The SPHINX1K model, obtained directly from the official repository, served as our pre-trained foundation model. The training was performed with a batch size of 4 and the learning rate was set to $2 \times 10{^{-5}}$.

\vspace{-0.05cm}
\subsection{Metrics}
\vspace{-0.05cm}
To quantitatively analyze the advantages of the \methodname~ method, we adhere to the experimental setup delineated in ~\cite{li2024learning} and have conducted the following metric calculations on the aRTF dataset: \textbf{Mean Average Precision (mAP)} is crucial for evaluating the accuracy of our model in detecting keypoints. We have calculated the mAP at three distinct L2 distance thresholds 2, 4, and 8 pixels from the ground truth keypoints. This approach allows us to measure the model's precision at various letolerance levelswhich is essential for understanding its robustness in different scenarios.
\textbf{Average Keypoint Distance (AKD)} serves as a direct measure of the geometric accuracy of the predicted keypoints relative to the ground truth. Leveraging the AKD is instrumental in assessing the model's spatial localization accuracy, thereby providing overall precision in keypoint detection.

\vspace{-0.05cm}
\subsection{Experiental Results}
\vspace{-0.05cm}
\subsubsection{Comparison on type-specific method}
In this experiment, we evaluate the performance of our SKT model and compare it against type-specific models under two different training settings: Sim-to-Real and Sim+Real-to-Real. The Sim-to-Real setting involves training the models solely on synthetic data, while the Sim+Real-to-Real setting incorporates both synthetic and real data during training. The results are presented in Table~\ref{tab:main-results}.

\textbf{Sim-to-Real:} Under the Sim-to-Real setting, the SKT model significantly outperforms the type-specific models across all garment types. The SKT model achieves an \APAll of 63.3 for T-shirts, 56.7 for shorts, and 83.9 for towels, notably surpassing the best-performing type-specific models. In contrast, the type-specific T-shirt, Shorts, and Towel Detectors achieve only 58.2, 51.4, and 83.2 in \APAll, respectively. More importantly, the SKT model delivers substantially better keypoint accuracy, with an AKD of 6.7 pixels for T-shirts, 2.9 pixels for shorts, and 3.4 pixels for towels, while the type-specific models exhibit much higher error rates (e.g., 14.0 pixels for T-shirts and 27.3 pixels for shorts).

\begin{figure}[t]
    \centering
    \includegraphics[width=0.95\linewidth]{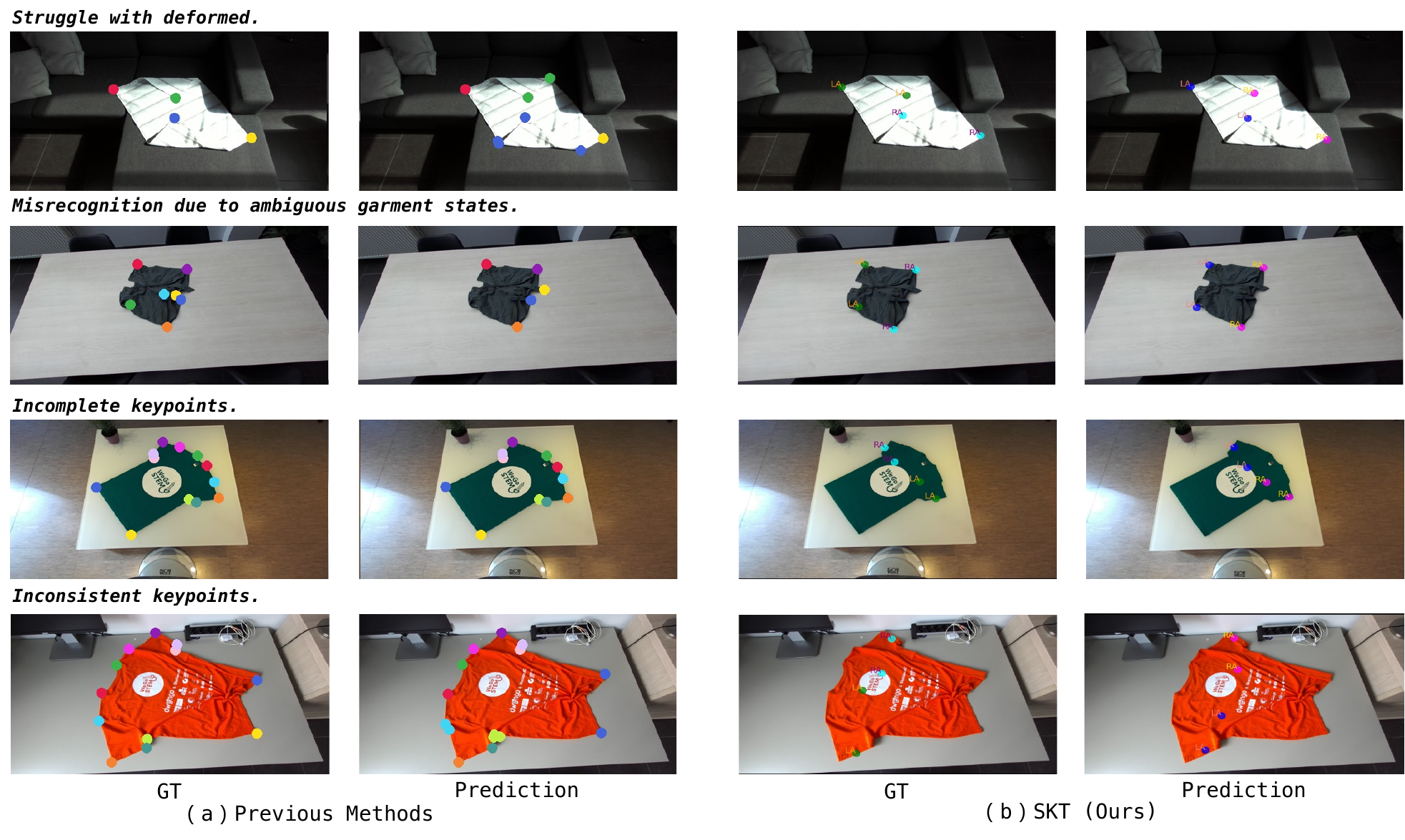}
    \caption{Qualitative visual comparison. \textbf{(a)} The previous approach~\cite{lips2024learning} struggles to handle deformed or ambiguous garment states, often resulting in incomplete and inconsistent keypoint predictions. \textbf{(b)} In contrast, our method provides more robust and accurate keypoint detection across diverse garment configurations, as demonstrated through improved visualization and performance.}
    \label{Fig: qualify-1}
\vspace{-4mm}
\end{figure}

These results highlight the strength of our unified approach. By learning across different garment types, the SKT model generalizes better than the one-to-one type-specific models, which struggle with the complexity and variability in garment configurations. The superior performance of the SKT model under the Sim-to-Real setting demonstrates its ability to generalize well without requiring specific models for each garment type, effectively overcoming the limitations of type-specific learning.

\textbf{Sim+Real-to-Real:} Incorporating real-world data in training further enhances SKT's performance. SKT generates action tuple keypoints and trajectories, achieving \APAll scores of 66.8, 59.9, and 86.8 for T-shirts, shorts, and towels respectively, with improved keypoint accuracies of 8.1, 7.1, and 3.0 pixels. SKT\_KP, which only predicts keypoints (similar to type-specific methods), achieves \APAll scores of 71.7, 60.3, and 88.2, with keypoint accuracies improved by 6.8, 3.9, and 2.0 pixels for the same garment types.

Our unified architecture enables transferable learning across garments, performing robustly even with complex deformations or ambiguous states. This single-model approach simplifies structure and training while outperforming type-specific models. It also mitigates overfitting risks, enhancing generalization across varied garment configurations and tasks. SKP\_KP results indicate that current metrics primarily reflect keypoint detection and that action tuple detection is more challenging than pure keypoint detection.

% \vspace{-0.02cm}
\subsubsection{Qualitative Analysis}
\vspace{-0.05cm}
In Fig.~\ref{Fig: qualify-1}, we analyze common failure cases of type-specific models during clothes manipulation tasks. These models often struggle with deformed or ambiguous garment states, such as highly crumpled or partially occluded clothing, resulting in incomplete, inconsistent keypoint predictions. For instance, when folds or complex garment shapes are present, type-specific models frequently misidentify semantic locations, generating false positives and misaligned keypoints. In the first row of Fig.~\ref{Fig: qualify-1}, type-specific models confuse key semantic locations in highly folded garments, while the second row illustrates further failures caused by overlapping fabric layers or ambiguous shapes, leading to significant errors in keypoint detection.

In contrast, our SKT model (b) offers a more robust and reliable solution by leveraging a fine-tuned vision-language model for general clothes manipulation. This allows the SKT model to handle diverse garment configurations with more consistent and accurate keypoint predictions. The visualizations in Fig.~\ref{Fig: qualify-1} clearly demonstrate the SKT model's ability to cope with challenging garment deformations and ambiguous states, resulting in fewer errors and more precise keypoint detection. The improved robustness of our model is also reflected in quantitative performance, where the SKT model significantly reduces false positives and enhances detection accuracy in scenarios where type-specific models struggle.

\vspace{-0.05cm}
\subsection{Ablation Study}
\vspace{-0.05cm}
To analyze SKT's design contributions, we conducted ablation studies following the original Sim-To-Real setting. Specifically, we examined three variations: 1)Replacing the "Any-Resolution" method from SPHINX~\cite{gao2024sphinx} with a default 224×224 image resolution for visual input. 2)Eliminating key-point detection tasks during training, focusing solely on direct trajectory tuple learning. 3)Implementing one-stage training instead of two-stage training.

Table~\ref{tab:ablation} illustrates that without high-resolution image input, model performance significantly deteriorates. We attribute this to the necessity of detailed visual information for cloth manipulation, as manipulation targets typically comprise only a small portion of the garment. Omitting key-point detection tasks also leads to suboptimal performance. We posit that these tasks serve as a bridge between the final action representation and pretrained general VQA tasks, facilitating better learning of action representation. Moreover, key-point detection tasks increase the volume of training samples, benefiting MLLM model training. Lastly, abandoning the staged training strategy results in performance degradation, aligning with previous findings that staged training aids MLLMs in adapting to new task formats.

\begin{table}[]
\centering
\vspace{-0.1cm}
\caption{Ablation study}
\vspace{-0.1cm}
\resizebox{0.95\columnwidth}{!}{%
\renewcommand\arraystretch{1.3}
\begin{tabular}{ccc|cc} 
\hline
High-Res & KP Task & Staged Training & \textbf{\APAllArrow}  & \textbf{\AKDArrow}  \\ 
\hline
\xmark        & \xmark       & \xmark               &  55.7     &  12.2     \\
\checkmark        & \xmark       & \xmark               & 64.7 &   9.8   \\
\checkmark        & \checkmark       & \xmark               & 65.3 &  8.9    \\
\checkmark        & \checkmark       & \checkmark               & 67.9 &  7.7    \\ 
\hline
\end{tabular}
\vspace{-0.1cm}
}
\label{tab:ablation}
\end{table}
\vspace{-0.1cm}

\vspace{-0.05cm}
\subsection{Discussions}
\vspace{-0.05cm}
We manually collected additional data that is not available in the existing dataset to evaluate the model's robustness, including samples with long pants, various folding states with deformations, and long sleeves for simple testing, as shown in the figure. The results indicate that our proposed method demonstrates relatively robust performance in handling pants and garments with different folding states. Notably, the model's ability to generalize from short pants to long pants is quite impressive, showcasing its potential for cross-category inference. However, it struggles with long sleeves, particularly folded long sleeves, which were not present in the training data. It indicates the need for further exploration of the model’s generalization capabilities, especially for unseen or diverse garment types and deformations.
\begin{figure}[t]
    \centering
    \includegraphics[width=0.95\linewidth]{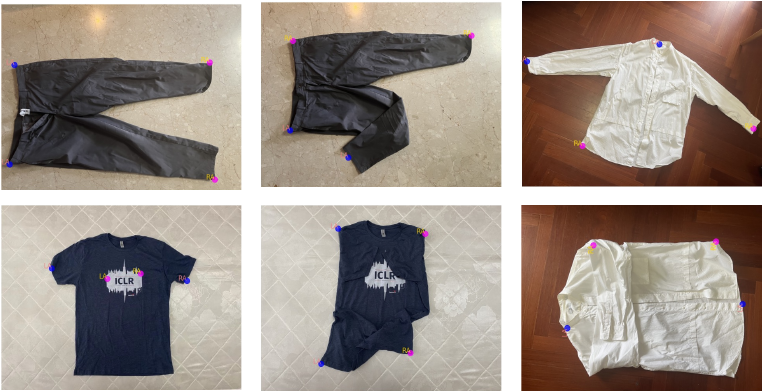}
    \caption{Evaluation of the SKT's performance on manually collected unseen data, including long pants, various folding states with deformations, and long sleeves. The results demonstrate robust handling of long pants and folded garments, while challenges remain in generalizing to unseen long sleeves and complex folded sleeve configurations.}
    \label{Fig: qualifynew}
\vspace{-4mm}
\end{figure}

\vspace{-0.1cm}
\section{Conclusion}
\vspace{-0.1cm}
This paper introduces a novel approach \methodname~to robotic garment manipulation using vision-language models, addressing the challenge of handling diverse and deformable garments with a unified framework. We propose a state-aware paired keypoint trajectory formulation that enhances the generalization across various garment states, including flat, folded, and deformed configurations. Additionally, we created a large-scale synthetic dataset with diverse garment states, significantly improving scalability by reducing dependence on real-world data. Experiments demonstrate that the integration of reasoning-based vision-language models enhances the robot's adaptability in complex scenarios for garment manipulation.

% \input{text/7_acknowledgments}

% \clearpage
{\small
\bibliographystyle{IEEEtranN}
\bibliography{ref}
}

\end{document}